\documentclass[preprints,article,accept,pdftex,moreauthors]{Definitions/mdpi} 
\firstpage{1} 
\pubvolume{1}
\issuenum{1}
\articlenumber{0}
\pubyear{2022}
\copyrightyear{2022}
\datereceived{} 
\dateaccepted{} 
\datepublished{} 
\hreflink{https://doi.org/} % If needed use \linebreak
\usepackage{array}
\newcolumntype{Y}{>{\centering\arraybackslash}X}
\pdfoutput=1

\Title{Automatic Assessment of Functional Movement Screening Exercises with Deep Learning Architectures}

\TitleCitation{Automatic Assessment of Functional Movement Screening Exercises with Deep Learning Architectures.}

\Author{Andreas Spilz $^{1,\dagger}$\orcidA{} and  Michael Munz $^{2,\dagger, *}$ \orcidB{}}

\AuthorNames{Andreas Spilz and Michael Munz}

\AuthorCitation{Spilz, A.; Munz, M.}
\address{%
$^{1}$ \quad Research Group Biomechatronics, University of Applied Sciences Ulm, 89081, Germany; Andreas.Spilz@thu.de\\
$^{2}$ \quad Research Group Biomechatronics, University of Applied Sciences Ulm, 89081, Germany; Michael.Munz@thu.de
}

\corres{Correspondence: Michael.Munz@thu.de}

\abstract{(1) Background: The success of physiotherapy depends on the regular and correct performance of movement exercises. A system that automatically evaluates these could support the therapy. Previous approaches in this area rarely rely on Deep Learning methods and do not yet fully use their potential. (2) Methods: Using a measurement system consisting of 17 IMUs, a dataset of four Functional Movement Screening (FMS) exercises is recorded. Exercise execution is evaluated by physiotherapists using the FMS criteria. This dataset is used to train a neural network that assigns the correct FMS score to an exercise repetition. We use an architecture consisting of CNN, LSTM and dense layers. Based on this framework, we apply various methods to optimize the performance of the network. For the optimization, we perform an extensive hyperparameter optimization. In addition, we are comparing different CNN structures that have been specifically adapted for use with IMU data. To test the developed approach, it is trained on the data from different FMS exercises and the performance is compared on unknown data from known and unknown subjects. (3) Results: The evaluation shows that the presented approach is able to classify unknown repetitions correctly. However, the trained network is yet unable to achieve consistent performance on the data of previously unknown subjects. Additionally, it can be seen that the performance of the network differs depending on the exercise it is trained for. (4) Conclusions: The present work shows that the presented deep learning approach is capable of performing complex motion analytic tasks based on IMU data. The observed performance degradation on the data of unknown subjects is comparably to publications of other research groups that relied on classical machine learning methods. However, the presented approach can rely on transfer learning methods, which allow to retrain the classifier by means of a few repetitions of an unknown subject. Transfer learning methods could also be used to compensate for performance differences between exercises.}


\begin{document}

%%%%%%%%%%%%%%%%%%%%%%%%%%%%%%%%%%%%%%%%%%
%\endnote{This is an endnote.} % To use endnotes, please un-comment \printendnotes below (before References). Only journal Laws uses \footnote.

% The order of the section titles is: Introduction, Materials and Methods, Results, Discussion, Conclusions for these journals: aerospace,algorithms,antibodies,antioxidants,atmosphere,axioms,biomedicines,carbon,crystals,designs,diagnostics,environments,fermentation,fluids,forests,fractalfract,informatics,information,inventions,jfmk,jrfm,lubricants,neonatalscreening,neuroglia,particles,pharmaceutics,polymers,processes,technologies,viruses,vision

\section{Introduction}

For successful physiotherapeutic treatment, the regular and correct execution of movement exercises at home is very important. A typical physiotherapy treatment begins with a detailed diagnosis, followed by initial treatment steps such as manual therapy. The patient is then taught various stretching and strengthening exercises. These exercises must be performed by the patient at home between appointments to ensure that the therapy progresses. However, training at home comes with certain problems. The movements performed here cannot be controlled and corrected by trained personnel. Thus, the patient can permanently perform the prescribed movements incorrectly, which on the one hand slows down the success of the treatment, and on the other hand can even aggravate the existing injury or add additional ones. Of course, it is not possible that every home training session is supervised by professionals. However, the work of physiotherapists could be significantly supported by a new technological development that monitors the home training of patients. 
Funded by the project MyPhysio@Home we want to address this research topic. Based on the data of a portable measurement system, machine learning algorithms will evaluate a performed movement exercise. In the first step, these algorithms are to classify the movement quality on a scale. In a future step, the patient should additionally receive feedback that informs him about the errors in his execution. In this publication we would like to present the developments made for the first step. 

We want to explore this topic, using screening exercises from Functional Movement Screening (FMS) \cite{Cook.2014, Cook2.2014}. FMS is an assessment system consisting of seven exercises that can be used to systematically determine movement restrictions or weaknesses in the human musculoskeletal system. Each exercise is assigned a score of 3 (perfect execution), 2 (complete execution with compensation movements), 1 (incomplete execution, even with compensation movements) and 0 (pain occured). For each exercise there is a well-defined list of movement characteristics that must be fulfilled in order to receive a certain score. FMS is chosen because it is widely used and one of the most popular screening systems in the field of sports physiotherapy. In addition, Cook's system also convinced with high interrater and intrarater reliability values \cite{Bonazza.2017}. This circumstance is especially interesting for machine learning applications that depend on unambiguous label information. Hart et al. and Ordonez et al. have made suggestions for a suitable dataset based on existing approaches to automated exercise evaluation \cite{Hart.2021, O'reilly.review}. The dataset should contain recordings from an appreciably large number of subjects. It should contain many different variations of exercise execution both incorrect and correct. These variants should not be staged. To test the applicability of a methodology to multiple exercise types, the dataset should contain different exercises. Based on these recommendations, we recorded, processed, and labeled an FMS dataset with 20 subjects of 4 different exercises.

Studies in the field of exercise evaluation rely on different measurement systems. Depth cameras \cite{Uccheddu.2019, Parisi.2015, Parisi.2016, Luo.2020, Liao.2020} or RGB cameras in combination with human pose estimation \cite{Karashchuk.2021, Desmarais.2021} are often used. In order to track the entire body posture, several cameras are required at different positions in the training room. The training room requires a comparatively large empty area for this. In addition, these cameras must be calibrated to a common coordinate system, before a measurement is taken, in order to provide usable results. This process is time-consuming and requires considerable computer resources and experience. These demands on training room, prior knowledge, and resources cannot be expected of a user. 
A suitable alternative to these camera-based systems are inertial measurement units (IMUs). These units record acceleration, angular velocity and magnetic field strength on three axis. Attached to the body segments, these measuring units can be used to record the kinematic movement of a human being. In addition, they are comparatively inexpensive, do not require additional space and do not cause additional work for the user during operation in an appropriately designed measurement system. 

Based on this IMU data we want to automate the evaluation of FMS exercises. Current studies mainly rely on classical machine learning methods such as decision tree \cite{O'reilly.2017, Kianifar.2016, Chen.2013}, random forest \cite{Luo.2020, Whelan.2017} or support vector machines \cite{Kianifar.2016, Bevilacqua.2018}. In contrast, very few studies e.g. \cite{Lee.2020} to date use deep learning methods in combination with IMU data and exercise evaluation, as also noted in systematic reviews on this topic \cite{O'reilly.review, Hart.2021}. This is particularly interesting, as these techniques are already widely used in related topics. Human Activity Recognition (HAR), for example, has been performed with IMU data and deep learning methods for quite some time \cite{Hammerla.2016}. Deep Learning methods offer some advantages over classical Machine Learning methods, such as automatic feature engineering and the option to apply Transfer Learning methods. Ordonez et al. have shown that using a combined structure of CNNs and LSTMs it is possible to distinguish between different activities using IMU data \cite{Ordonez.2016}. 
Lee et al. have already compared the performance of a random forest approach with an approach based on a combination of a CNN and a LSTM on an exercise evaluation task \cite{Lee.2020}. When classifying a squat into different performance variants based on the data of five IMUs, the deep learning approach (accuracy: 91.7\%) achieved significantly better results than the classical approach (accuracy: 75.4\%).

 We will explore this approach further, adapt it to our measurement setup and the FMS exercises and look in particular at the following issues. Hart et al. already noted in their review of various exercise evaluation publications that hyperparameter optimization is often not performed, or not documented, even though the benefits are widely recognized \cite{Hart.2021}. Therefore, to obtain the best possible results, we performed a detailed optimization to determine the exact parameters for the best possible performance. We also present a network structure that allows repetitions of different lengths to be used for training without distorting the temporal context of the individual repetition. 
 
 We also want to test alternative CNN structures explicitly designed to process IMU data. CNNs are originally designed for the analysis of information in image form. By arranging the individual IMU channels as rows of a two-dimensional matrix, CNNs can also analyze this time-varying information. Nevertheless, the two types of information differ significantly. In the case of an image, CNNs perform spatial convolution, whereas in the case of IMU data, convolution occurs over time, the individual measured quantities (acceleration, angular velocity, etc.), and the spatial relationship between IMU units. We want to investigate whether the performance of a classifier improves when this fact is taken into account by an adapted CNN structure. Recent studies \cite{Muenzner.2017, Grzeszick.2017, Rueda.2018, Rueda.2021} already investigate the influence of different CNN structures on classification performance, we want to adapt this approaches to a new network structure and apply it in exercise evaluation.
In summary, the main contributions of this paper are as follows: 

\begin{itemize}
    \item Introduction of a novel IMU-based dataset for the automatic evaluation of FMS exercises
    \item Development of a neural network based approach for the automatic evaluation of FMS exercises
    \item Evaluation of the influence of different hyperparameters and network structures on classification performance.
    \item Performance evaluation of the developed system on different exercises from the FMS
\end{itemize}

%%%%%%%%%%%%%%%%%%%%%%%%%%%%%%%%%%%%%%%%%%
\section{Materials and Methods}
\subsection{Dataset}
In the context of a study, we create a labeled dataset of four different FMS exercises. The exact specification of this dataset and our methodology is explained in the following.
\subsubsection{Exercises}
\begin{figure}[h]
    \centering
    \includegraphics[width=13 cm]{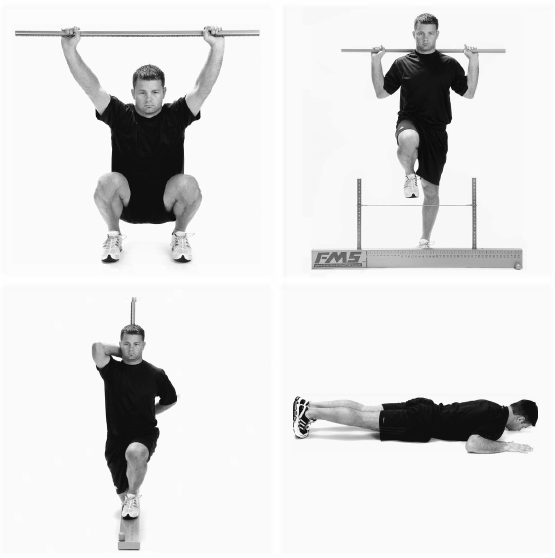}
    \caption{FMS exercises included in the recorded dataset. From left to right, from top to bottom: Deep Squat (DS), Hurdle Step (HS), Inline Lunge (IL) and Trunk Stability Pushup (TSP). \cite{Cook.book}}
    \label{fig:fms_exercises}
\end{figure}

The described dataset contains four exercises from the FMS. Specifically, these are the Deep Squat (DS), Hurdle Step (HS), Inline Lunge (IL), and Trunk Stability Pushup (TSP) exercises. These are modified versions of the commonly known sports exercises squat, pushup and lunge. The HS movement is rather unknown. Here, a subject stands upright in front of a hurdle, the height of which depends on the person's anatomy. Now the subject must lift one leg over this hurdle and touch the ground on the other side with his heel. A sample image of each exercise can be found in figure \ref{fig:fms_exercises}. Additional information on the exact movement sequence of each exercise can be found in Cook et al. \cite{Cook.book}.

\subsubsection{Study population}
The exercises presented are performed by a total of 17 healthy volunteers. During recruitment, care is taken to ensure that approximately equal numbers of men (9) and women (11) participate in the study. The age of the volunteers ranges from 24 to 62 years, trying to recruit as balanced as possible from the different age groups (41 \textpm 13.53 years). The height of the subjects ranges from 160 to 185 cm (173.5 \textpm 8.3 cm) and the weight ranges from 40 to 110 kg (71.35 \textpm 14.9 kg). We try to include athletic, average and rather inactive subjects in the study population. Subjects self-assess their level of athleticism.
Before the trial begins, each participant signs an informed consent form. The study protocol was evaluated and accepted by the ethics committee of the University of Applied Sciences, Ulm. The study presented is registered in the German Clinical Trials Register and can be found under the ID DRKS00027259.
\subsubsection{Measurement setup}
\begin{figure}[h]
    \centering
    \includegraphics[width=10 cm]{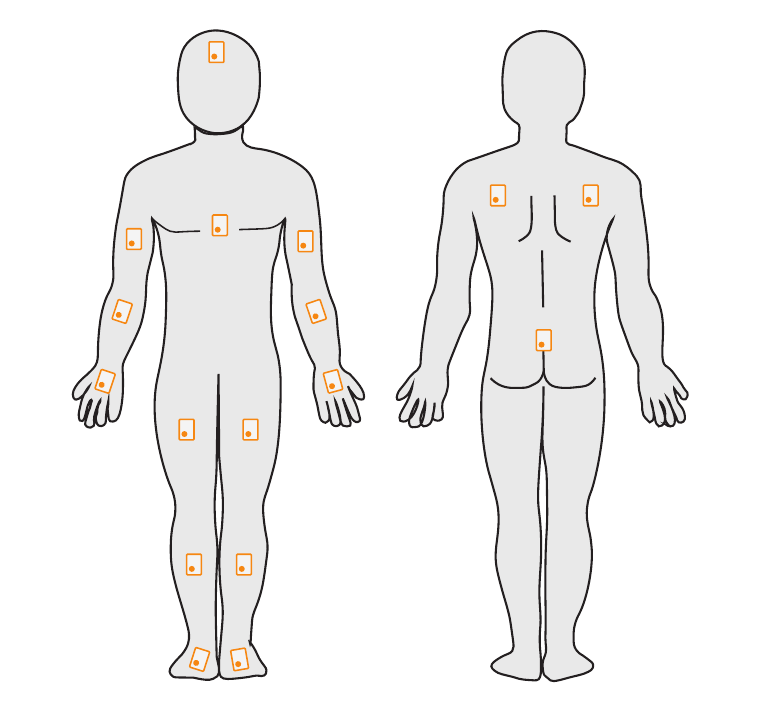}
    \caption{Positioning of the IMUs (orange outline) on the subject. }
    \label{fig:imu_positions}
\end{figure}

Each subject is instrumented with a total of 17 IMUs. We have divided the human body into 17 segments, to each of which an IMU is attached. The number and position of the individual IMUs is based on the \emph{Xsens MVN fullbody} tracking system \cite{XsensMVN}. The setup definition of \emph{Xsens} seems suitable to us, because \emph{Xsens} has years of experience in professional motion analysis with IMUs and the products are already used in a wide variety of sports. In addition, the products are already used by other research groups in many areas of motion analysis. The exact positioning of the IMUs is shown in figure \ref{fig:imu_positions}. The individual IMUs are attached to the subject with elastic straps. The position on a segment is chosen in a manner that minimizes displacement of the IMUs while in motion, e.g. by placing them between muscle bellies.
In the described setup we use \emph{Shimmer3} IMUs by \emph{ShimmerSensing}. These IMUs feature a three-axial accelerometer, gyrometer and magnetometer and an air pressure sensor. The individual metrics are recorded with a sampling frequency of 120 Hz on a measuring range of \textpm 16 g (accelerometer), \textpm 200 °/s (gyrometer), \textpm 49.5 Ga (magnetometer) and 300-1100 hPa (air pressure sensor).
To allow for later evaluation of the FMS exercises, the subject is filmed from five perspectives (left, right, top, front and back) with RGB cameras. The \emph{Basler acA640-120gc} is used for this purpose, with a resolution of 640 x 480 px and a sampling rate of 120 Hz.

\subsubsection{Procedure}
After their arrival, the subjects are informed about the procedure. Subsequently, the IMUs are attached to the previously defined locations. After all measurement systems have been started, a synchronization procedure is performed (for details see \ref{section:postprocessing}) and the actual recording starts. The subjects now have to complete a total of three rounds of an exercise circuit. Each round consists of 15 repetitions of DS, 15 repetitions of HS with the right foot, 15 repetitions of HS with the left foot, 15 repetitions of TSP, 15 repetitions of IL with the right foot in front and 15 repetitions with the left foot in front. The 15 repetitions are again divided into five units of 3 repetitions each. In between each unit, there is a few seconds of rest. After each completed unit, there is a few seconds rest and a longer break after each completed round. Since the overall program is very strenuous, the subjects had the option to stop the experiment at any point. After the experiment is completed, another synchronization procedure is performed.

\subsubsection{Labeling}
The individual repetitions are evaluated by three physiotherapists on the basis of the video recordings, according to the FMS criteria. However, only the ratings 1 (incomplete execution, even with compensation movements), 2 (complete execution with compensation movements) and 3 (perfect execution) are assigned, since the rating 0 is assigned when pain occurs. This adjustment is made because pain is difficult to detect without input from the subject, and also because the study protocol placed great emphasis on stopping immediately if pain occurred.  

Several measures have been implemented to guarantee reliable labels: 
\begin{itemize}
\item	the three physiotherapists are trained uniformly and have equal experience with FMS
\item	in a test phase, 100 repetitions are labeled by each physiotherapist, subsequently inconsistencies are discussed and corrected.
\item	each repetition is rated by three physiotherapists, the ratings are given independently of each other
\item habit effects are avoided by randomly selecting the sample to be rated from all repetitions
\end{itemize}

To optimize the labeling process, a web application is developed that presents the different video perspectives in an organized way and plays them. The application allows the rater to directly enter his feedback for the present repetition and proceed to the next repetition. 

After labeling was completed, interrater reliability was assessed. For this purpose, Krippendorff's alpha \cite{Krippendorff.2008} is used as an assessment metric. This metric is suitable in the present case because, unlike Cohen's Kappa for example, it can be used for more than two raters. Moreover, it is suitable for ordinal data, such as the ratings of the FMS.

\subsubsection{Postprocessing}
\label{section:postprocessing}
After the measurement, the collected data from the cameras and the IMUs is time-synchronized. The cameras were already synchronized during the measurement, with a common trigger signal from a synchronization box. The IMUs are synchronized with each other and with the cameras after the measurement. Subsequently, the collected data from the cameras and the IMUs are time-synchronized. The cameras are already synchronized during the measurement, as they can be triggered by a square wave signal. Each camera received the same trigger signal from a synchronization box. The IMUs are synchronized with each other and with the cameras after the measurement. Since the high number of IMUs is very difficult to synchronize by normal means, an alternative method was developed. This method is based on a precisely defined pulsing magnetic field to which the IMUs are exposed before and after the measurement. With the method described and evaluated in Spilz et al. the individual IMUs can thereby be synchronized with each other with an accuracy of below 2.6 ms \cite{Spilz.2021}. In addition, the magnetic field is controlled by the same box that is responsible for the synchronization of the cameras. Via this box, the two synchronization methods can be put into a temporal context that allows the cameras and the IMUs to be synchronized. 

\begin{figure}[h]
    \centering
    \includegraphics[width=10 cm]{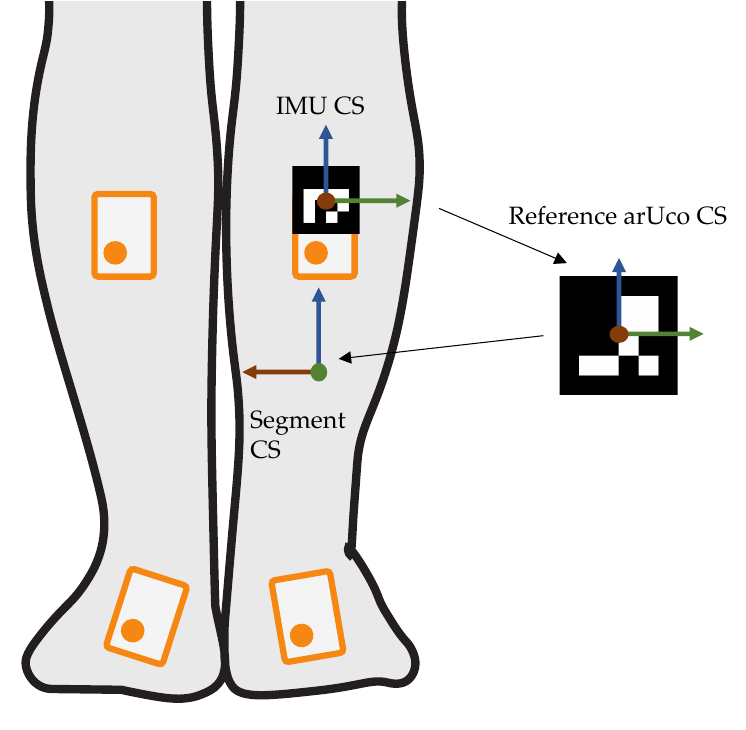}
    \caption{Derivation of the transformation between the IMU CS and the body segment CS. The CS consist of x- (red), y- (green) and z-axis (blue).}
    \label{fig:aruco}
\end{figure}

Measures are taken to ensure that the IMUs are both attached to the correct body segment and have a consistent orientation to the segment. For this purpose, a so-called arUco marker \cite{aruco} is attached to each IMU. These markers have a unique ID that can be retrieved from a 2D RGB image. Furthermore, the orientation of the markers relative to the camera can be determined. We attach the IMUs to the subject and then photograph him or her from several perspectives. These images can be used to verify that the individual IMUs are located on the correct segment and also to determine the orientation of the IMUs relative to the body segment. The orientation determination is performed in the following steps: The orientation of the IMU coordinate system (CS) relative to the camera CS is determined via the arUco on the IMUs. In the background of the image, reference arUco markers are placed on the floor and on the wall. These can be used to determine the orientation of the room CS to the camera CS. The subject is clearly positioned in a well-defined posture relative to this room CS. Thus, the orientation of the individual segments to the room CS is known and the orientation of the IMU CS to the body segment CS can be determined (see figure \ref{fig:aruco}. The resulting transformation is applied to the three-dimensional IMU data, so the measured quantities are in the same segment CS for each subject. 

\subsection{Architecture and Training}
In the presented work, we want to introduce a method to assign an FMS rating of 1, 2 or 3 to an exercise repetition. This classification task will be performed by a neural network. The structure and the hyperparameters of this network are presented in the following.
Our approach is based on a neural network that consists of CNN layers, LSTM layers and dense layers. The input is first passed to the CNN layers, whose task it is to learn relevant features. These features are passed on to the LSTM layers, which analyze temporal dependencies. Finally, the dense layers follow, which are responsible for the actual classification. In the following, we would like to develop the exact structure.

\subsubsection{Variable sequence length}
\label{section:vsl}
To process the time series data from the IMUs with the CNN layers, we follow the approach of Ordonez et al. and arrange the accelerometer and gyrometer IMU channels as rows of a 2D matrix \cite{Ordonez.2016}. The channels are sorted IMU by IMU, so that the x, y and z axes of the accelerometer come first, then those of the gyrometer and then the accelerometer channels of the next IMU. 

In HAR, the common approach is to divide these 2D matrices into windows of equal size. Each of these windows is then used as a sample for the dataset. This procedure cannot be adopted for exercise evaluation. The reason is the labels: In HAR, an unambiguous label can still be assigned to a window. Even if the window contains e.g. only a fraction of a step, the performed activity and therefore the assigned label, is still "walking". This approach is not applicable to exercise evaluation, since the existing labels, such as the FMS evaluations, always refer to the entire repetition. If a repetition is labeled as 2 (i.e. errors occurred during the execution), it is not clear which windows of the exercise is responsible for this label. One windows might contains the perfect partial execution of a squat, while another window contains the incorrect partial execution responsible for the assinged label 2. Since the labels only indicate whether errors occur during the execution and not when, the individual windows cannot be provided with unique labels. Accordingly, an entire repetition must always be made available to the network as a sample. But this approach introduces a different issue.                   
The duration of the individual repetitions varies significantly. Typically problems like this are addressed by resampling the repetitions to a uniform number of time steps \cite{Lee.2020, Hart.2021}. The disadvantage of this approach is that the temporal context is lost. However, this context is elementary when IMU data is used, e.g. the velocity of the IMU depends on the measured acceleration relative to measured time. Therefore, we want to preserve the temporal context and pass it to the neural network. To achieve this, all repetitions are first zero-padded to the same length. Then each repetition is divided into X equal-sized windows without overlap (see figure \ref{fig:nn_input}). This 3D representation (IMU channels x time steps x windows) is then used as input to the neural network. The CNN layers are located within the \emph{TimeDistributed} wrapper of \emph{Tensorflow} 2.5 \cite{tensorflow2015-whitepaper}. This wrapper allows that several windows of an input are processed by one and the same layer. This results in the weights of the filters being learned based on all windows, but the windows are still processed individually. Following these CNN layers, the windows are then transferred as steps to the LSTM layer. The LSTM is preceded by a masking layer, which ensures that the windows that contain only 0's are skipped. Since the LSTM layer can handle any number of time steps, it is possible to process a repetition of any length. With this, the general architecture of the network is defined, now we want to identify the best possible setup.

\subsubsection{Hyperparameter optimization}
\label{section:ho}
To define the exact parameters of the network structure, we perform a hyperparameter optimization.
Table \ref{table:1} lists the parameters that were varied, along with the range in which they were varied. The parameters were chosen based on our own experience for the presented network structure and supplemented according to the recommendations of Yang et al. \cite{Yang.Hyper.2020}. 
Enclosed are explanations of the terms used: 
\begin{itemize}[leftmargin=*,labelsep=5.8mm]
    \item activation function: the activation function for each layer except the LSTM layers and the output layer
    \item CNN-block: refers to a block consisting of a CNN layer, followed by a max-pooling layer and either a dropout or a batch normalization layer
    \item combination scheme for CNN-blocks: the individual CNN-blocks have different hyperparameters depending on the chosen scheme. 
    \item "increasing filters, fixed kernel size": means the number of filters increases per CNN-block [16, 32, 64] but the kernel size remains the same [(5x5), (5x5), (5x5)]. 
    \item "increasing filters, decreasing kernel size": means the number of filters increases per CNN-block [16, 32, 64] and the kernel size is decreased [(9x9), (5x5), (3x3)].
    
\end{itemize}

\begin{table}[t]
\centering
\renewcommand{\arraystretch}{1.5}
\begin{tabularx}{\textwidth}{||Y Y||} 
 \hline
 Parameter & Range \\ [1ex]
 \hline\hline
 Activation function & ReLU / ELU / LReLU  \\ [1ex]
 \hline
 Number of CNN-blocks & 1 / 2 / 3 \\ [1ex]
 \hline
 Combination scheme for CNN-blocks &  increasing filters, fixed kernel size / increasing filters, decreasing kernel size  \\ [1ex]
 \hline
 Regularization technique &  dropout (rate 0.2) / batch normalization  \\ [1ex]
 \hline
 Number of LSTM layers & 1 / 2 \\ [1ex] 
 \hline
 Batch size & 4 / 8 / 16 / 32 \\ [1ex]
 \hline
\end{tabularx}
\caption{Table of the parameters that were varied for the hyperparameter optimization, along with the range in which they were varied.}
\label{table:1}
\end{table}

\begin{figure}[t]
    \centering
    \includegraphics[width=10 cm]{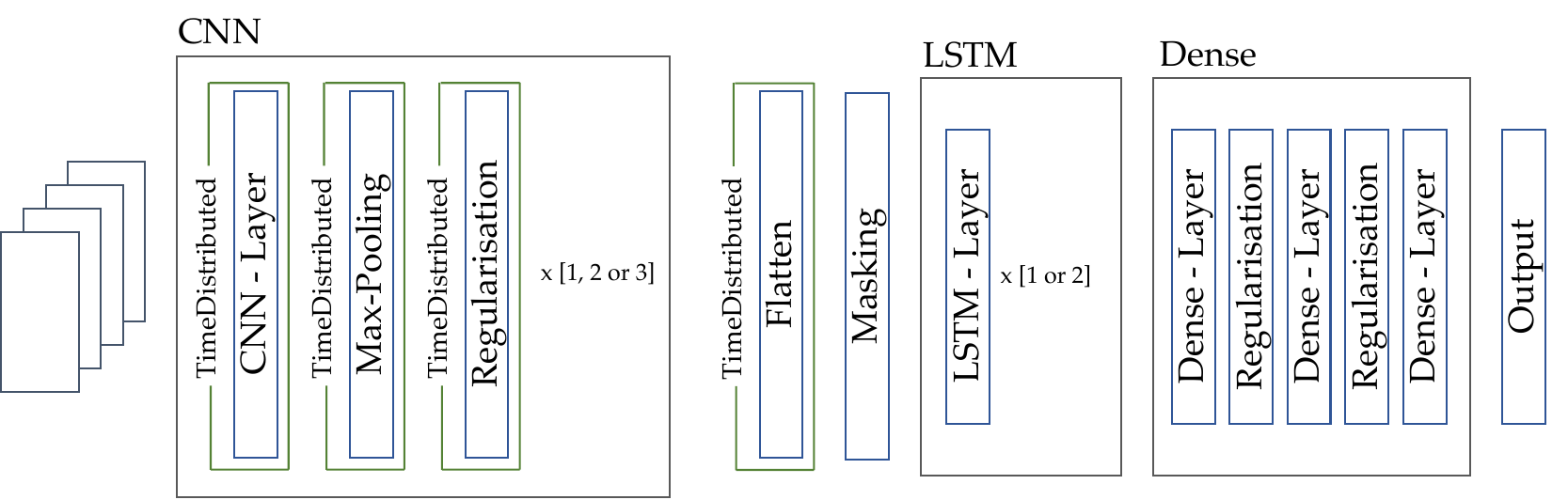}
    \caption{Network architecture of the CNN-LSTM}
    \label{fig:network}
\end{figure}

The remaining hyperparameters are defined as follows. The number of units in every LSTM layer is 256. The number of neurons in the first dense layer is 512, 128 in the second and 3 in the last (output) layer. In the output layer a softmax function is used as activation function, in the LSTM layers we use tanh as activation function and the sigmoid function for the recurrent activation, in all other layers the function that performs best in the optimization is used. The max-pooling operation is performed with a (2x2) kernel. Every trainable weight is initialized using the uniform initialization by Glorot. As evaluation metric we use the macro F1-score \cite{f1macro}. This is the arithmetic mean of the f1-scores of the individual classes. As error function we use categorical cross-entropy. Adam with a learning rate of 0.0001 is used as optimizer.
This optimization is executed using the DS repetitions of the presented dataset. In order to get an objective impression of the performance, each parameter combination is tested in a 5-fold cross validation (CV). The dataset is divided into training (74\%), validation (16\%) and test (20\%) set using a stratified shuffle split for each fold. The datasets are stratified according to the label distribution of the entire dataset. The described network is implemented in Tensorflow 2.5. The resulting network structure is shown in figure \ref{fig:network}.

\begin{figure}[b]
    \centering
    \includegraphics[width=10 cm]{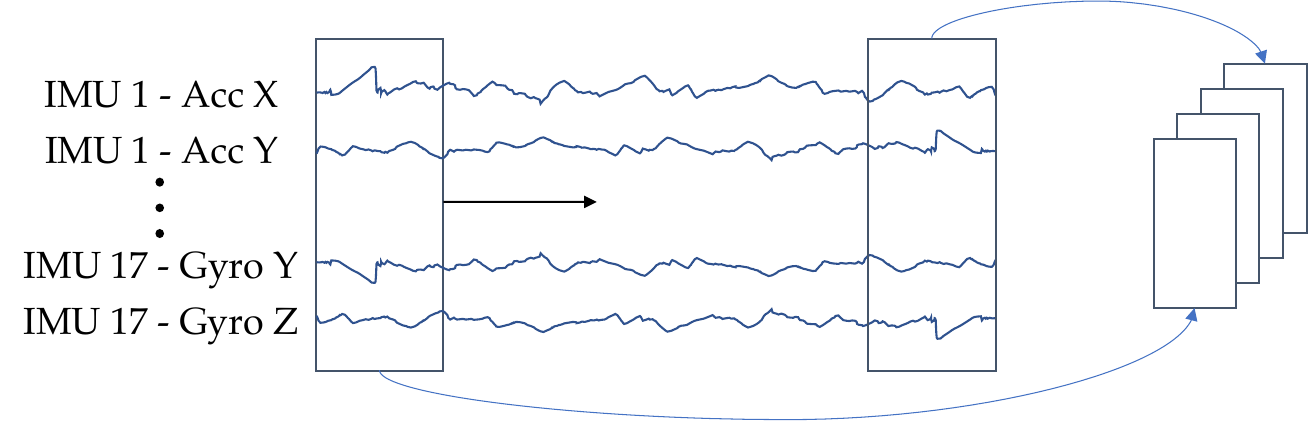}
    \caption{Input representation for the CNN-LSTM. The individual sensor channels are combined line by line so that they result in an image-like representation.}
    \label{fig:nn_input}
\end{figure}

The channels of the IMUs are arranged and windowed according to figure \ref{fig:nn_input} to generate a 2D representation. The data was standardized, but separately for accelerometer and gyrometer data. 

A total of 90 different parameter combinations is evaluated. For each of these runs, a value for each parameter in table \ref{table:1} is randomly chosen. The selection is performed with a \emph{WandB sweeps agent} \cite{wandb} operating in random mode.

\subsubsection{Alternative CNN structures}
\label{section:imu_specific}
The previously used approach is based on arranging IMU data in two dimensions and applying a two-dimensional convolution to it. There also exist alternative CNN approaches that are specifically designed for IMU data. We want to adapt two existing approaches to our network structure and compare the performance with the previously described "baseline" approach. So three different structures are tested:

- "baseline" approach: The approach described in section \ref{section:vsl}

\begin{figure}[t]
    \centering
    \includegraphics[width=10 cm]{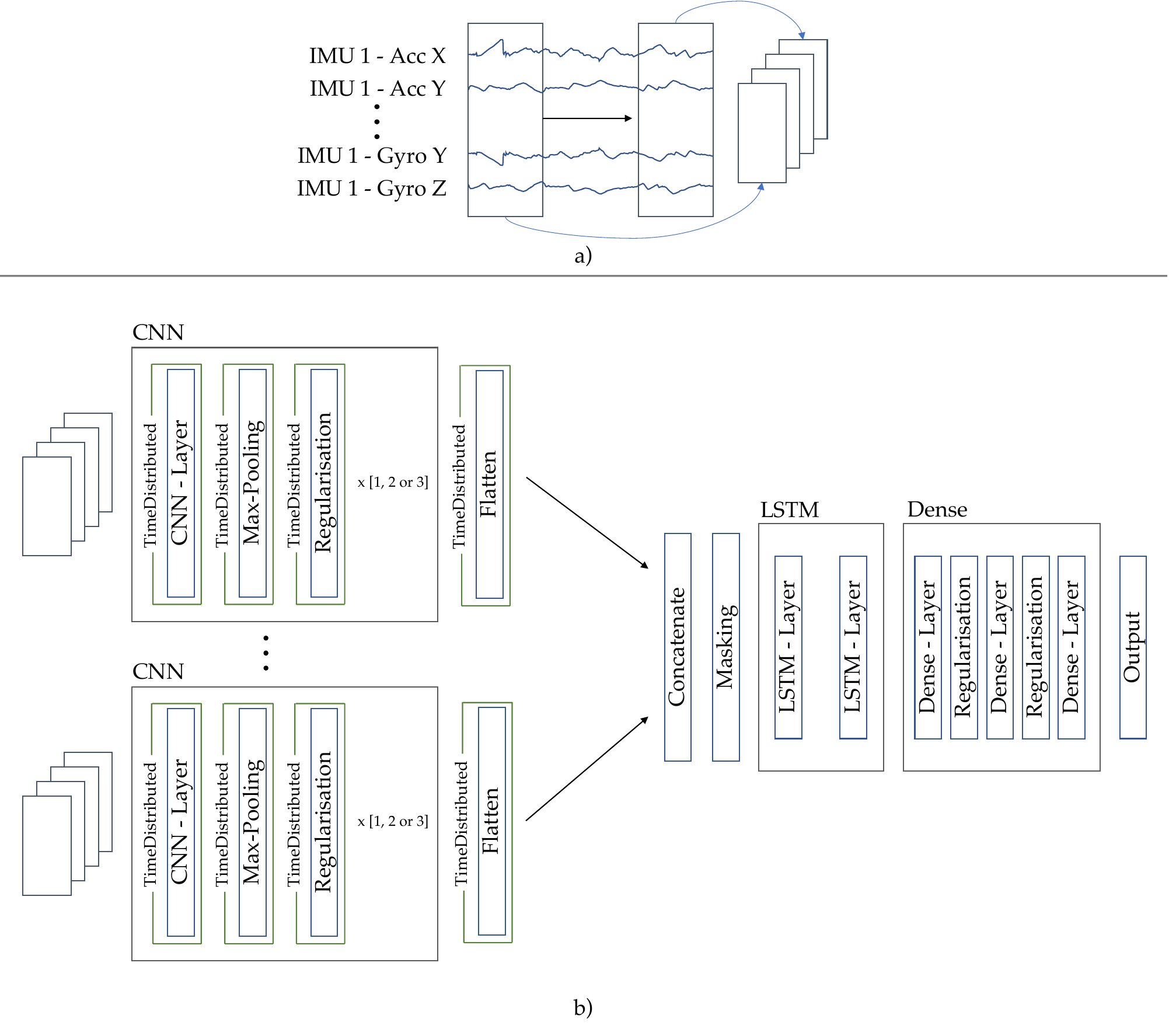}
    \caption{a) Input definition for the tested "IMU-centric" approach. b) Network architecture of the "IMU-centric" approach }
    \label{fig:imu_specific}
\end{figure}

- "IMU-centric" approach: Following Grzeszick et al. the data of all IMUs are no longer combined to a 2D representation and processed, but the data of each IMU is processed in a separate network branch \cite{Grzeszick.2017, Rueda.2018}. With this approach, the data from the individual IMUs is merged at a later stage and individual filters are learned for each IMU. We are adapting our previous procedure to follow the same approach. For this, the data is zero-padded and windowed as before, but now separated by IMUs (see figure \ref{fig:imu_specific} a)). The network structure is modified, so that the input data from each IMU is processed by its own network branch, which consists of the CNN-blocks already described. Within these blocks, a channel's data is convoluted over the time axis, using kernels with the size (1x5). Afterwards, the network outputs are concatenated and passed to a masking layer followed by an LSTM layer. From here on, the network structure is identical to the previously described architecture (see figure \ref{fig:imu_specific} b)). 

- "channel-centric" approach: 
Following Münzner et al, the IMU data are convoluted along the time axis channel by channel, but all with the same kernel \cite{Muenzner.2017}. Due to the temporal convolution, the data of the individual channels are not mixed. The shared kernel also ensures that the same features are scanned for on all channels. To adapt this approach, little is changed in the existing framework from chapter \ref{section:vsl}. Only the kernel size is adapted (1x5), the preprocessing of the data and the network structure remain the same.  This approach ensures that the channels are convoluted along the time axis, with the learned filter being the same for each channel. 

The modified CNN structures may lead to a different feature learning process. This process may now require additional or fewer CNN layers to learn meaningful features. So the different approaches can reach their full potential, we will perform an appropriate series of experiments. Each of the three approaches will be tested with one, two or three CNN-blocks. As in chapter \ref{section:vsl}, these blocks consist of a CNN layer, a max-pooling layer and a dropout layer. As the number of CNN-blocks increases, the kernel size in each block remains the same, while the number of filters doubles (16 / 32 / 64). The kernel of the max-pooling layer always remains the same (2x2), as well as the rate of the dropout layer (rate=0.2). In total, 9 different configurations are tested. A 5-fold cross-validation is performed with each configuration, with a training (74 \%) validation (16\%) and test (20 \%) split, stratified by labels. This optimization is executed using the DS repetitions of the presented dataset. 

The remaining hyperparameters are chosen on the basis of the results of chapter \ref{section:ho}. 
Accordingly, both LSTM layers have 256 units, and the dense layers have 512, 128 and 3 neurons. In the output layer a softmax function is used as activation function, in the LSTM layers we use tanh as activation function and the sigmoid function for the recurrent activation, in all other layers we use ELU. Every trainable weight was initialized using the uniform initialization by Glorot. As evaluation metric we use the macro F1-score. As error function we use categorical cross-entropy. Adam with a learning rate of 0.0001 is used as optimizer and the batch size is 32.

\subsubsection{Evaluation of the performance on different exercises}
\label{section:different_exercises}
To this point, all analyses have been performed with the DS repetitions. In the next step, we want to examine whether the developed approach can also be applied to the three other exercises from the presented dataset. For this we use the approach that has performed best so far, i.e. the baseline approach from chapter \ref{section:imu_specific}, with three CNN-blocks. The remaining hyperparameters are also adopted from this chapter. To get an objective overview of the performance, the network is trained with a 5-fold cross validation, as specified in \ref{section:ho}. Since the HS and IL exercises can each be performed with an active left or right foot, different variants are considered here: First, for each of the two exercises, one network is trained with the repetitions of the left side only and one network is trained with the repetitions of the right side only. In addition, a network with all repetitions of the left and right side is trained to investigate how the performance of the network is affected. 

\subsubsection{Evaluation of the performance on unknown subjects}
Finally, we would like to investigate how well the developed approach performs if repetitions of unknown subjects are classified. For this investigation, we utilize the same network configuration as in chapter \ref{section:different_exercises}. However, this time a Leave-One-Subject-Out (LOSO) split is performed, so the test set consists of the data of one subject, which is not part of the training and validation set. The remaining data is divided into training (80\% of remaining data) and validation (20\% of remaining data) set. This process is performed 10 times, each time with a different subject and is called Leave-One-Subject-Out-Cross-Validation (LOSOCV). 

This analysis is performed individually for each of the four exercises in the data set, whereby HS and IL are again split in three variants: left, right, and combined. In contrast to previous analyses, we use the weighted F1-score instead of the macro F1-score. The reason for this is the label distribution within the data of a single subject. Many subjects received the same score for every performed repetition of a particular exercise. Since these examples represent the entire test data set in the performed LOSOCV, the macro F1-score would distort the performance evaluation.
%%%%%%%%%%%%%%%%%%%%%%%%%%%%%%%%%%%%%%%%%%
\section{Results}

\subsection{Dataset}
\begin{figure}[t]
    \centering
    \includegraphics[width=10 cm]{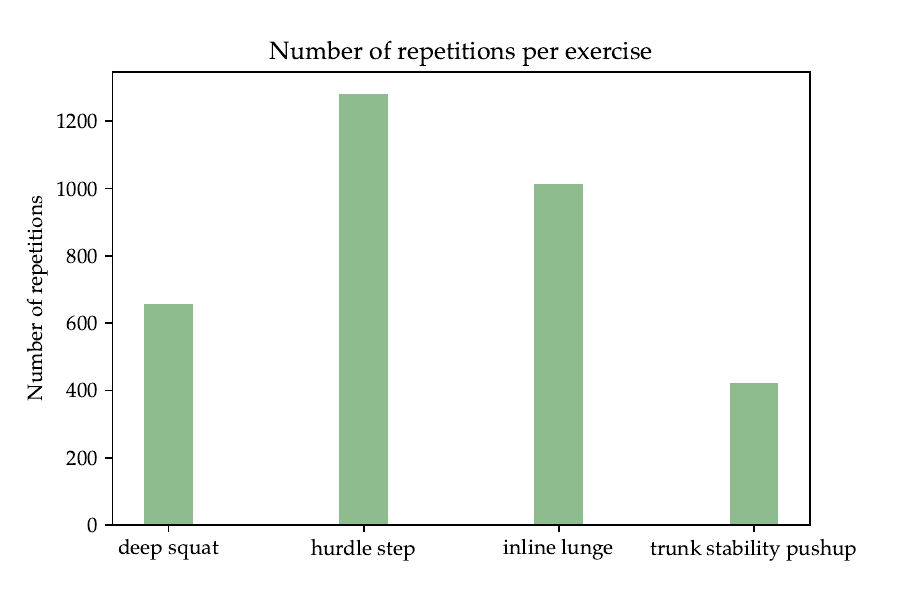}
    \caption{Number ob repetitions per FMS exercise in the presented dataset }
    \label{fig:reps_per_exercise}
\end{figure}

The described study resulted in a dataset with 3374 useable repetitions. The exact number of repetitions per exercise is shown in figure \ref{fig:reps_per_exercise}. Not all subjects were physically able to perform the intended number of repetitions, so the experiment was terminated when the subject was no longer able to continue without injury. In particular, the TSP could only be performed correctly by a few subjects due to the physically demanding exercise format. The HS and IL exercises are present more frequently than the other two. The reason for this is, that the exercises were theoretically performed 90 times (45 left, 45 right) per subject, while the other exercises were performed just 45 times.

To ensure that the labels assigned are reliable, each recurrence is labeled by three physical therapists who independently submitted their ratings. Krippendorff's alpha for the ratings given is 0.815 for the entire dataset, indicating a high level of agreement among the individual raters \cite{krippendorff.book}.

\begin{figure}[t]
    \centering
    \includegraphics[width=13 cm]{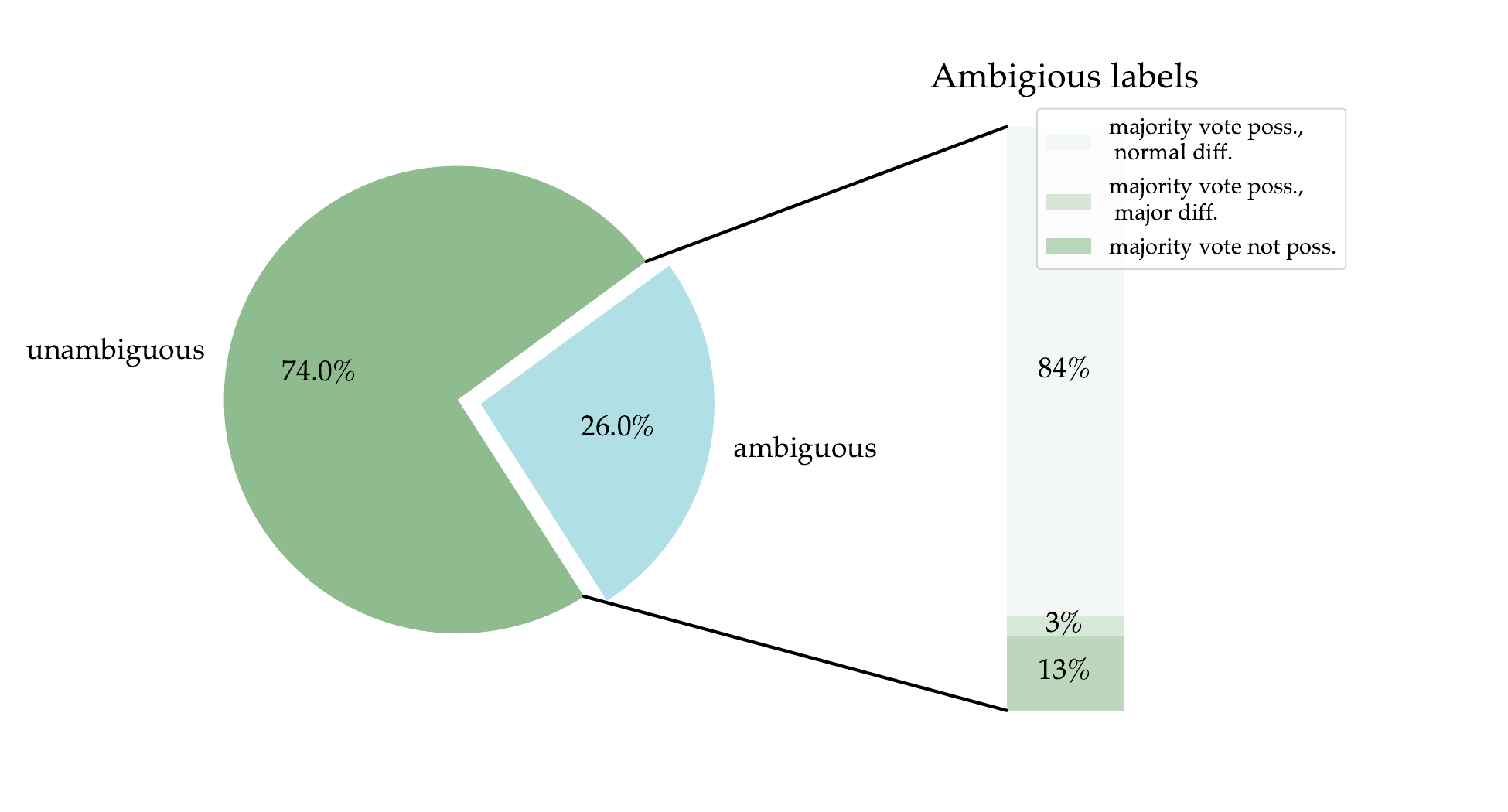}
    \caption{Listing of the different degrees of agreement in the evaluation and their share in the total number of repetitions. For "unambiguous" repetitions, all raters agree. For "ambiguous" repetitions, at least one does not agree with the rest. The category "ambiguous" can be further divided into repetitions where no majority vote is possible, repetitions where a majority vote is possible, but there are major differences in the given ratings and repetitions where a majority vote is possible, but just minor differences in the given ratings.}
    \label{fig:label_distribution}
\end{figure}

The final evaluation of a repetition is determined by a majority vote. Figure \ref{fig:label_distribution} shows a visualization of the proportions of repetitions that have a certain degree of agreement. The largest proportion of repetitions (74~\%) is unambiguous, i.e. all raters agree. The remaining 26~\% of the repetitions show inconsistencies in the rating. This proportion can be further subdivided into repetitions where a majority vote is possible and the discrepant score differs by only one point, and repetitions where a majority vote is possible but the discrepant score differs by more than one point. Additionally, there is a category for repetitions where a majority decision is no longer possible, because all given scores differ from each other. These repetitions are excluded from the dataset for the time being. In total, 3374 repetitions are usable for the development of the algorithms. 

\begin{figure}[t]
    \centering
    \includegraphics[width=13 cm]{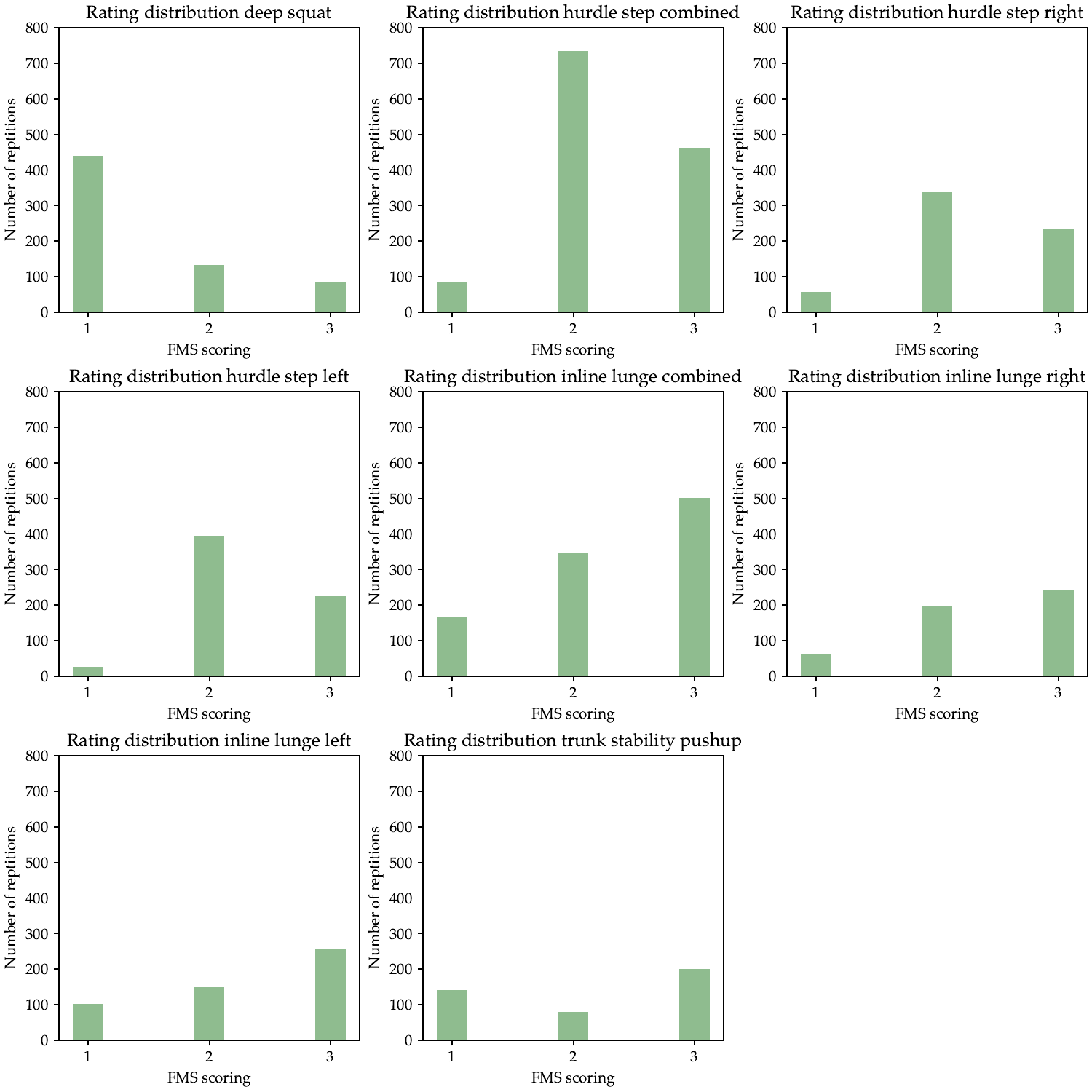}
    \caption{Distribution of the assigned FMS ratings (1, 2 or 3) divided by exercise. HS and IL can be performed with an active left or right foot and are therefore split into, e.g. "HS right" and "HS left". The distribution over both left and right repetitions is presented with the label "combined", e.g. HS combined }
    \label{fig:label_distribution_per_exercise}
\end{figure}

The label distribution for each exercise is shown in figure \ref{fig:label_distribution_per_exercise}. As can be seen, the label distribution for each exercise is skewed. Interestingly, a different rating is over represented for each exercise. The clearest differences are found in the HS left, where the rating 2 is about 14 times more frequent than the rating 1. The most balanced distribution is found for the TSP, here the ratings 1 and 3 occur only 1.5 times and 2 times as frequently as the rating 2.

\subsection{Hyperparameter optimization}
\label{section: resHyp}
The optimal parameter combination is selected using the averaged performance from the 5-fold CV. Considered is the averaged macro F1-score achieved on the training, validation and test dataset. 
The best performance was achieved by a network with the following hyperparameters: batch size 32, three CNN-blocks with the "increasing filter, fixed kernel size" scheme, dropout with a rate of 0.2 as regularization technique and two LSTM layers. This combination results in a macro F1-score of 0.96 on the training set, 0.95 on the validation set and 0.9 on the test set. In all tested configurations, macro F1-scores were achieved in the ranges of 0.93 - 1 (training set), 0.94 - 0.96 (validation set) and 0.87 - 0.9 (test set).

\subsection{Alternative CNN structures}

Table \ref{table:2} lists the performance metrics of the different CNN architectures. The mean value of the macro F1-scores of 5-fold CV is shown here. The performance differences between the three approaches used are minimal. Regardless of the number of CNN blocks and the CNN structure used, the macro F1-score on training (0.94 - 0.97), validation (0.88 - 0.9) and test (0.95 - 0.96) set is found to be close to the optimum. The performance on the different partial data sets is also very balanced; no significant overfitting can be observed in any of the 9 configurations. 

\begin{table}[t]
\renewcommand{\arraystretch}{1.5}
\begin{tabularx}{\textwidth}{||Y Y Y Y||}
 \hline
 CNN-blocks  & IMU-specific (train / validation / test) & channel-specific (train / validation / test) & baseline (train / validation / test) \\ [0.5ex] 
 \hline\hline
1 & 0.95 / 0.89 / 0.95 & 0.96 / 0.9 / 0.96  & 0.94 / 0.89 / 0.96   \\ 
2 & 0.96 / 0.9 / 0.96 & 0.94 / 0.9 / 0.95 & 0.97 / 0.88 / 0.96 \\
3 & 0.95 / 0.9 / 0.96   & 0.94 / 0.88 / 0.95  & 0.95 / 0.9 / 0.95 \\

 \hline
\end{tabularx}
\caption{Results of the evaluation of different IMU-specific architectures. Displayed metrics are the arithmetic mean of the macro F1-scores achieved on the training / validation / test dataset. }
\label{table:2}
\end{table}

\subsection{Evaluation of the performance on different exercises}
\label{section: evDiffEx}
The results of the performance evaluation on different datasets are summarized in table \ref{table:3}. For each dataset, the mean and standard deviation of the macro F1-score from the 5-fold CV are shown.

Several observations can be made using this data: The mean performance on the training, validation and test data is comparable for both DS and TSP, and by far the best in comparison. The comparison between the different sets indicates a slight overfitting on the training data. 

The performance of IL combined, IL left and IL right does not vary considerably between each other and no overfitting can be noted. However the performance is significantly worse than the performance of TSP and DS. 

Compared to the other exercises, the performance of the HS variants is the poorest. In addition, HS combined, HS left and HS right differ significantly in the performance levels achieved. However, the three variants show no evidence of overfitting. 

The calculated standard deviation of the performance metrics is very low for all considered datasets.
\begin{table}[t]
\renewcommand{\arraystretch}{1.5}
\begin{tabularx}{\textwidth}{||Y Y Y Y||}
 \hline
 dataset & training set & validation set & test set \\
  \hline\hline
 Hurdle Step  & 0.69 $\pm$ 0.05 & 0.68 $\pm$ 0.05 & 0.65 $\pm$ 0.05 \\
 Hurdle Step right  & 0.73 $\pm$ 0.04 & 0.76 $\pm$ 0.06 & 0.69 $\pm$ 0.04 \\
 Hurdle Step  left & 0.58 $\pm$ 0.04 & 0.57 $\pm$ 0.02 & 0.55 $\pm$ 0.02 \\
 Inline Lunge & 0.88 $\pm$ 0.04 & 0.86 $\pm$ 0.05 & 0.83 $\pm$ 0.02 \\
 Inline Lunge right & 0.86 $\pm$ 0.04 & 0.82 $\pm$ 0.06 & 0.84 $\pm$ 0.04 \\
 Inline Lunge left & 0.87 $\pm$ 0.01 & 0.85 $\pm$ 0.05 & 0.85 $\pm$ 0.05 \\
 Trunk Stability Pushup & 0.95 $\pm$ 0.03 & 0.9 $\pm$ 0.04 & 0.91 $\pm$ 0.04 \\
 Deep Squat & 0.94 $\pm$ 0.03 & 0.95 $\pm$ 0.01 & 0.9 $\pm$ 0.02 \\
 \hline
\end{tabularx}
\caption{Results of the evaluation of the proposed approach on different datasets. For each dataset a 5-fold CV is performed and the arithmetic mean of the macro F1-scores of training, validation and test set is shown, together with the corresponding standard deviation.}
\label{table:3}
\end{table}

\subsection{Evaluation of the performance on unknown subjects}
\label{section: resEvUnSubj}
Table \ref{table:4} shows the results of the performance test on the data of unknown subjects. A 10-fold LOSOCV was performed for each exercise, and the mean and standard deviation of these 10 runs is shown. 

The achieved performance on training and validation dataset is comparable to the one reported in chapter \ref{section: evDiffEx}. The exact numbers differ from each other since the weighted F1-score is used in this analysis instead of the macro F1-score. Otherwise, the results are similar: overfitting is only observable to a small extent and the best performance is achieved on the DS and TSP data sets. IL combined, IL right and IL left achieve again similar, but poorer results than TSP and DS. HS combined, HS right and HS left achieve more homogeneous results than before and also reach a significantly higher value.

Striking is the performance on the data of the unknown subject, i.e. on the test set. On average, poor to very poor results are obtained on all data sets, which also scatter significantly. 

\begin{table}[t]
\renewcommand{\arraystretch}{1.5}
\begin{tabularx}{\textwidth}{||Y Y Y Y||}
 \hline
 dataset & training set & validation set & test set \\
  \hline\hline
 Hurdle Step  & 0.81 $\pm$ 0.02 & 0.82 $\pm$ 0.02 & 0.3 $\pm$ 0.28 \\
 Hurdle Step right  & 0.84 $\pm$ 0.04 & 0.79 $\pm$ 0.02 & 0.27 $\pm$ 0.26 \\
 Hurdle Step  left & 0.82 $\pm$ 0.02 & 0.87 $\pm$ 0.02 & 0.41 $\pm$ 0.41 \\
 Inline Lunge & 0.91 $\pm$ 0.03 & 0.88 $\pm$ 0.03 & 0.33 $\pm$ 0.18 \\
 Inline Lunge right & 0.86 $\pm$ 0.03 & 0.81 $\pm$ 0.02 & 0.44 $\pm$ 0.35 \\
 Inline Lunge left & 0.88 $\pm$ 0.03 & 0.81 $\pm$ 0.04 & 0.5 $\pm$ 0.35 \\
 Trunk Stability Pushup & 0.95 $\pm$ 0.02 & 0.95 $\pm$ 0.01 & 0.15 $\pm$ 0.32 \\
 Deep Squat & 0.98 $\pm$ 0.01 & 0.94 $\pm$ 0.01 & 0.49 $\pm$ 0.43 \\
 \hline
\end{tabularx}
\caption{Results of the evaluation of the proposed approach on unknown subjects. For each dataset a 10-fold LOSOCV is performed and the arithmetic mean of the weighted F1-scores of training, validation and test set is shown, together with the corresponding standard deviation.}
\label{table:4}
\end{table}

%%%%%%%%%%%%%%%%%%%%%%%%%%%%%%%%%%%%%%%%%%
\section{Discussion}

\subsection{Measurement setup}
Since we want to develop a universally applicable approach for exercise evaluation in our research, we chose a versatile IMU setup with a large number of sensors. This is the opposite approach of many other research groups \cite{Hart.2021}, who aim to achieve results with a setup as minimal as possible. These minimal setups offer the advantage, that they can be used immediately in practice without causing an unreasonable amount of additional work for the user. However, we think it is reasonable to acquire data with a maximized configuration. This way, the influence of the number of IMUs used on the performance of the evaluation algorithm can be investigated in a feature analysis. These findings in turn allow it to develop a reduced setup for the final application.

\subsection{Neural network architecture}
Our approach to using arbitrarily long repetitions was developed based on the fact that the currently available labels are assigned to the whole repetition. If this information were available, one could follow the approach from HAR and divide a repetition into multiple windows, each with its own label. Accordingly, one could work without zero-padding and time-distributed wrappers and also have a larger number of shorter samples. However, the question arises whether a network based on an excerpt is still able to evaluate the entire repetition. The reason for this is the scoring schemes, which differ from exercise to exercise. There are, for example, incorrect movement sequences that only lead to a certain evaluation if the occur in combination. If the network has only one of these segments at its disposal, it does not have the possibility to evaluate such combined criteria.

\subsection{Hyperparameter Optimization}
During the optimization, we have already compared the most common parameters and identified the best possible combination for the application at hand. Yet still it should be considered that this optimization was carried out on the basis of one exercise and a limited subject collective. 
Accordingly, no statement can yet be made as to whether this parameter combination is also optimal for other exercises. In addition, not every possible parameter combination was checked, but 90 random combinations. The identified configuration thus corresponds to the tendencies that could be read from this subset of possibilities. Decisions for the parameters, as well as for the dataset and the number of combinations tested, were made with feasibility in mind. In order to keep the total computing time within limits, but still obtain as much information as possible, the setup chosen represented a viable middle ground.

\subsection{Alternative CNN structures}
 We compared three different approaches to process the IMU data within the CNN component of the neural network. Overall, no obvious difference in the performance of the three approaches was found. 
 The approaches tested were selected based on the results of Münzner and Rueda et al. \cite{Muenzner.2017, Rueda.2018}. However, it should be checked whether other approaches lead to an improvement of the performance on the present problem. For example, there is a sensor-specific approach, which is constructed in such a way that separate filters are learned for the accelerometer and gyrometer data. This setup could potentially lead to improved performance and should be investigated.

\subsection{Evaluation of the performance on different exercises}

Depending on which dataset is used, there are significant differences in performance. This behavior will be examined in more detail below. First, it is noticeable that the results of chapter \ref{section: evDiffEx} and \ref{section: resEvUnSubj} differ significantly, although the performance on training and validation dataset should be similar. The reason for this are the used evaluation metrics and the class distribution (see figure \ref{fig:label_distribution}) of the individual datasets. For example, the HS variants achieved a macro F1-score of 0.58 - 0.72 on the training dataset, but a weighted F1-score of 0.82 - 0.86. Based on the rating distribution of these variants, one can see that rating 1 is underrepresented and accordingly a misclassified example of rating 1 influences the macro F1-score significantly more than misclassified examples of the other classes. The influence of a misclassified example on the weighted F1-score is independent of rating. Especially in the case of the HS variants and the IL variants, one can conclude from the described discrepancy between the metrics that there are significant differences in the performance on the different ratings. The reason is probably the insufficient number of training examples for certain ratings, which can be seen in figure \ref{fig:label_distribution}. A good example is the macro F1-score of the HS left dataset: This is 0.58 on the training dataset, at the same time there are hardly any examples for rating 1. 

Another reason for the performance discrepancies could be the differences between the exercises. Each exercise has a completely different movement sequence and different evaluation criteria. Some of these criteria are certainly more complex to learn than others. Especially for the complex criteria, there may be too few examples with too little variation to learn a correct representation.

Action should be taken to align HS and IL performance with TSP and DS performance. To compensate for missing examples for certain classes, one could try to use transfer learning effects. The first processing steps for the analysis of the IMU data should be very similar for all exercises. Accordingly, one could train an appropriately constructed autoencoder with the data of all exercises and use the learned weights of the encoder for the CNN blocks. The higher number of available examples, which also cover a higher variability, could lead to more robust features and thus to increased performance in the evaluation of the different exercises.

\subsection{Evaluation of the performance on unknown subjects}
In this work, we used a LOSOCV split to test the performance of the networks on data from unseen subjects. This led to consistently poor results on the test sets, which also feature considerable variability. Other publications in the field of automatic exercise evaluation encounter similar problems \cite{Hart.2021, O'reilly.review}. Often, so-called global classifiers are trained on data from all subjects, which then perform reasonably well on a test set of data from those subjects. This approach was also used in the present publication in chapter X. In addition, "personal" classifiers are trained with the data of one specific subject and perform well on unseen repetitions from that subject. However, with the present Deep Learning structure, one could also go a different way. Instead of two separate classifiers, the already trained global classifier could also be expanded with an additional, previously unknown person. With a transfer learning approach, the network could be retrained with a few repetitions of the additional person. Other application areas that use transfer learning methods achieve very good results with significantly less training effort and data. Accordingly, an already trained "global" classifier could be extended by a new person with less training effort and training examples than necessary for the training of a "personal" classifier.

\subsection{Segmentation}
 For the intended application, another factor should be considered. In this publication, exercise repetitions were used that were extracted by a human. For the final application, this segmentation must be automated. An appropriate algorithm must detect and extract the individual exercises from a stream of measurement data. This automated segmentation could change the characteristics of the data. For example, it would be possible that the automatically segmented repetitions are cut out very precisely, while the previously used repetitions contain some inactivity at the beginning and end. Perhaps the trained network pays special attention to this inactivity and as a result the performance on automated repetitions deteriorates significantly. This possibility should be avoided and therefore the network should be trained again with repetitions extracted by an automated method.

%%%%%%%%%%%%%%%%%%%%%%%%%%%%%%%%%%%%%%%%%%
\section{Conclusions}
In the presented work, a novel dataset containing approx. 3300 rated repetitions of various FMS exercises was presented. 

Based on this dataset, a network structure was developed that allows to classify repetitions into a certain quality category. This network structure is composed of CNN layers, LSTM layers and dense layers. By means of an optimization, different options for the hyperparameters were tested and the best combination was identified. 

In addition, the performance of different CNN approaches, designed specifically for the processing of IMU data, was evaluated. The baseline approach proved to be the most useful. Further investigations can be done in this area, e.g. a sensor-based approach can be tested, which processes the input data ordered by sensor (accelerometer, gyrometer). 

The final evaluation of the developed deep learning approach on various exercises shows that the approach is able to classify unknown repetitions from known individuals. However, the performance on unknown data from unknown participants is not sufficient. This problem is already known from other publications. In the present case we have the possibility to use transfer learning methods to address the problem. By means of a few repetitions of a previously unknown subject, it could be possible to adapt the classifier to additional subjects with little effort.

The evaluation also showed that the performance achieved varies depending on the exercise. Using transfer learning methods, it is likewise possible to develop a classifier that learns more general, robust features, which in turn should lead to more consistent performance.

%%%%%%%%%%%%%%%%%%%%%%%%%%%%%%%%%%%%%%%%%%

%%%%%%%%%%%%%%%%%%%%%%%%%%%%%%%%%%%%%%%%%%
\vspace{6pt} 

%%%%%%%%%%%%%%%%%%%%%%%%%%%%%%%%%%%%%%%%%%
%% optional
%\supplementary{The following supporting information can be downloaded at:  \linksupplementary{s1}, Figure S1: title; Table S1: title; Video S1: title.}


%%%%%%%%%%%%%%%%%%%%%%%%%%%%%%%%%%%%%%%%%%

\authorcontributions{conceptualization, A.S. and M.M.; methodology, A.S. and M.M.; software, A.S.; validation, A.S. and M.M; formal analysis, A.S. and M.M.; investigation, A.S.; resources, M.M; data curation, A.S.; writing--original draft preparation, A.S.; writing--review and editing, M.M.; visualization, A.S.; supervision, M.M.; project administration, M.M.; funding acquisition, M.M.}

%%%%%%%%%%%%%%%%%%%%%%%%%%%%%%%%%%%%%%%%%%
%\acknowledgments{In this section you can acknowledge any support given which is not covered by the author contribution or funding sections. This may include administrative and technical support, or donations in kind (e.g., materials used for experiments).}

%%%%%%%%%%%%%%%%%%%%%%%%%%%%%%%%%%%%%%%%%%
\conflictsofinterest{The authors declare no conflict of interest.}

%%%%%%%%%%%%%%%%%%%%%%%%%%%%%%%%%%%%%%%%%%

%% Only for journal Encyclopedia
%\entrylink{The Link to this entry published on the encyclopedia platform.}

%%%%%%%%%%%%%%%%%%%%%%%%%%%%%%%%%%%%%%%%%%
\begin{adjustwidth}{-\extralength}{0cm}

\reftitle{References}


%\bibliography{your_external_BibTeX_file}

%=====================================
% References, variant B: internal bibliography
%=====================================
\externalbibliography{yes}
\bibliography{references.bib}

% If authors have biography, please use the format below
%\section*{Short Biography of Authors}
%\bio
%{\raisebox{-0.35cm}{\includegraphics[width=3.5cm,height=5.3cm,clip,keepaspectratio]{Definitions/author1.pdf}}}
%{\textbf{Firstname Lastname} Biography of first author}
%
%\bio
%{\raisebox{-0.35cm}{\includegraphics[width=3.5cm,height=5.3cm,clip,keepaspectratio]{Definitions/author2.jpg}}}
%{\textbf{Firstname Lastname} Biography of second author}

% For the MDPI journals use author-date citation, please follow the formatting guidelines on http://www.mdpi.com/authors/references
% To cite two works by the same author: \citeauthor{ref-journal-1a} (\citeyear{ref-journal-1a}, \citeyear{ref-journal-1b}). This produces: Whittaker (1967, 1975)
% To cite two works by the same author with specific pages: \citeauthor{ref-journal-3a} (\citeyear{ref-journal-3a}, p. 328; \citeyear{ref-journal-3b}, p.475). This produces: Wong (1999, p. 328; 2000, p. 475)

%%%%%%%%%%%%%%%%%%%%%%%%%%%%%%%%%%%%%%%%%%
%% for journal Sci
%\reviewreports{\\
%Reviewer 1 comments and authors’ response\\
%Reviewer 2 comments and authors’ response\\
%Reviewer 3 comments and authors’ response
%}
%%%%%%%%%%%%%%%%%%%%%%%%%%%%%%%%%%%%%%%%%%
\end{adjustwidth}
\end{document}